%% file: main.tex
\documentclass[sigconf]{acmart}

\acmConference[ICSE 2022]{The 44th International Conference on Software Engineering}{May 21–29, 2022}{Pittsburgh, PA, USA}

\usepackage{algorithm}
\usepackage{enumitem}
\setlist[itemize]{align=parleft,left=1pt..1em}

\AtBeginDocument{%
  \providecommand\BibTeX{{%
    \normalfont B\kern-0.5em{\scshape i\kern-0.25em b}\kern-0.8em\TeX}}}

\copyrightyear{2022}
\acmYear{2022}
\setcopyright{acmlicensed}\acmConference[ICSE '22]{44th International Conference on Software Engineering}{May 21--29, 2022}{Pittsburgh, PA, USA}
\acmBooktitle{44th International Conference on Software Engineering (ICSE '22), May 21--29, 2022, Pittsburgh, PA, USA}
\acmPrice{15.00}
\acmDOI{10.1145/3510003.3510171}
\acmISBN{978-1-4503-9221-1/22/05}

\usepackage{algorithm}
\usepackage{algpseudocode}

\usepackage[utf8]{inputenc} 
\usepackage[T1]{fontenc}    
\usepackage{hyperref}       
\usepackage{url}            
\usepackage{booktabs}       
\usepackage{amsfonts}       
\usepackage{nicefrac}       
\usepackage{microtype}      
\usepackage{xcolor}         
\usepackage{listings}       
\usepackage{amsmath}
\usepackage{multirow}
\usepackage{tkz-kiviat} 
\usetikzlibrary{arrows}
\graphicspath{ {./figures/} }

\usepackage{pgfplots}
\pgfplotsset{compat=1.17}
\usepackage{caption}
\usepackage{subcaption}
\usepackage{indentfirst}
\usepackage{bm}
\input macros

\begin{document}

\title{DeepSTL - From English Requirements to Signal Temporal Logic}

\author{Jie He}
\email{jie.he@tuwien.ac.at}
\affiliation{%
  \institution{Technische Universität Wien}
  \city{Vienna}
  \country{Austria}
}

\author{Ezio Bartocci}
\email{ezio.bartocci@tuwien.ac.at}
\affiliation{%
  \institution{Technische Universität Wien}
  \city{Vienna}
  \country{Austria}
}

\author{Dejan Ničković}
\email{Dejan.Nickovic@ait.ac.at}
\affiliation{%
  \institution{AIT Austrian Institute of Technology}
  \city{Vienna}
  \country{Austria}
}

\author{Haris Isakovic}
\email{haris@vmars.tuwien.ac.at}
\affiliation{%
  \institution{Technische Universität Wien}
  \city{Vienna}
  \country{Austria}
}

\author{Radu Grosu}
\email{radu.grosu@tuwien.ac.at}
\affiliation{%
  \institution{Technische Universität Wien}
  \city{Vienna}
  \country{Austria}
}

\input abstract

\maketitle

\input{1_Introduction}
\input{2_related_work}
\input{3_STL}

\input{4_STL_distribution}

\input{5_corpus_construction}
\input{6_machine_translation}

\input{7_discussion}
\input{8_conclusion}

\begin{acks}
This  project  has  received  funding  from  the  European Union’s  Horizon  2020  research  and  innovation  programme  under  grant  agreement No 956123, and funding from the Austrian FFG ICT of the Future program under grant agreement No 880811.
\end{acks}

\bibliographystyle{unsrt}
\bibliography{main.bbl}

\end{document}

%% file: macros.tex
\newcommand{\until}{\textbf{U}}
\newcommand{\since}{\textbf{S}}
\newcommand{\F}{\textbf{F}}
\newcommand{\GG}{\textbf{G}}
\newcommand{\opP}{\textbf{O}}
\newcommand{\opH}{\textbf{H}}
\newcommand{\true}{\textbf{true}}
\newcommand{\false}{\textbf{false}}
\newcommand{\rise}{\textbf{rise}}
\newcommand{\fall}{\textbf{fall}}

\newcommand{\SP}{\text{SP}}
\newcommand{\TP}{\text{TP}}
\newcommand{\NTP}{\text{NTP}}
\newcommand{\PP}{\text{P}}

\newcommand{\UTO}{\textbf{UTO}}
\newcommand{\BTO}{\textbf{BTO}}

%% file: abstract.tex
\begin{abstract}
\setlength{\parindent}{0pt}
Formal methods provide very powerful tools and techniques for the 
design and analysis of complex systems. Their practical application 
remains however limited, due to the widely accepted belief that 
formal methods require extensive expertise and a steep learning 
curve. Writing correct formal specifications in form of logical 
formulas is still considered to be a difficult and error prone 
task.

In this paper we propose DeepSTL, a tool and technique for the 
translation of informal requirements, given as free English sentences, into 
Signal Temporal Logic (STL), a formal specification language for cyber-physical 
systems, used both by academia and advanced research labs in industry. A major 
challenge to devise such a translator is the lack of publicly available 
informal requirements and formal specifications. We propose a two-step workflow 
to address this challenge. We first design a grammar-based generation technique 
of synthetic data, where each output is a random STL formula and its associated 
set of possible English translations. In the second step, we use a state-of-
the-art transformer-based neural translation technique, to train an accurate 
attentional translator of English to STL.  The experimental results show high 
translation quality for patterns of English requirements that have been well 
trained, making this workflow promising to be extended for processing more 
complex translation tasks. 
\end{abstract}

%% file: 1_Introduction.tex
\section{Introduction}

``What is reasonable is real. That which is real is reasonable.''
This famous proposition from Hegel, saying that 
everything has its ``logic'', often resonates in 
Alice's mind. Alice is a verification engineer responsible 
for safety-critical cyber-physical systems (CPS). She 
advocates the use of formal methods with requirements specified 
in logic, as part of the development of complex CPS. 

Formal specifications enable rigorous reasoning about a CPS product (for example its model checking or systematic testing) during all its design phases, as well as during operation (for example via runtime verification)~\cite{bartocci2018specification}. Alice is frustrated by the resistance of her colleagues 
to adopt formal methods in their design methodology. 
She is aware that one major bottleneck in a wider acceptance of these techniques results from the steep learning curve to translate informal requirements expressed in natural language into formal specifications. The correspondence between a requirement written in English and its temporal logic formalization 
is not always straightforward, as illustrated in the example below:
\begin{itemize}
    \item \textbf{English Requirement}: 
    
    Whenever V\_Mot is detected to become equal to 0, then at a time point starting after at most 100 time units Spd\_Act shall continuously remain on 0 for at least 20 time units.
    \item \textbf{Signal Temporal Logic (STL)}: 
    
    $\GG (\rise (\text{V\_Mot} = 0) \rightarrow \F_{[0,100]} \GG_{[0,20]} \ (\text{Spd\_Act} = 0))$
\end{itemize}


Bob is Alice's colleague and an expert in machine learning. He 
introduces Alice to the tremendous achievements in 
natural language processing (NLP), demonstrated by 
applications such as Google Translate and DeepL. Alice is 
impressed by the quality of translations between natural languages. 
She realizes that NLP is a technology that can 
reduce the gap between engineers and formal methods, and significantly 
increase the acceptance of rigorous 
specifications. 

However, Alice also observes 
that this potential solution does not come without challenges. 
In order to build a translator from one spoken language to another, there is a huge amount of texts available in both languages that can be used for training and there is also a series of systematic translation solutions. In contrast, for translating CPS requirements given in natural language into formal specifications, there are two major challenges:

\begin{itemize}
    \item \textbf{Challenge 1}: Lack of available training data. The informal requirement documents are sparse and often not publicly available, and formal specifications are even sparser.
    \item \textbf{Challenge 2}: No mature solutions for translating English requirements into formal specifications, where special features of these two languages need to be considered. 
\end{itemize}

In this paper, as a first attempt to adopt NLP to tackle the above two challenges, we propose DeepSTL, a method and associated tool for the translation of CPS requirements 
given in relatively free English to Signal Temporal Logic (STL)~\cite{maler2013monitoring}, a formal 
specification language used by 
the CPS academia and industry. To develop DeepSTL we 
address the following five research questions (\textbf{RQ}), the solutions of which are also our main contributions.


\begin{description}
\item[RQ1:] What kind of empirical statistics of STL requirements, found in scientific literature, can guide data generation?

\item[RQ2:] How to generate synthetic examples of STL requirements consistently with the empirically collected statistics?
\end{description}

The first two research questions are related to \textbf{Challenge 1}. For \textbf{RQ1}, empirical STL statistics in literature and practice are analyzed in Section~\ref{sec:empirical_stl}. For \textbf{RQ2}, we design in Section~\ref{sec:corpus_construction}, a systematic grammar-based generation of synthetic data sets, consisting of pairs of STL formulae and their associated set of possible English translations.

\begin{description}
\item[RQ3:] How effective is DeepSTL in learning synthetic STL?
\item[RQ4:] How well does DeepSTL extrapolate to STL requirements found in scientific literature?
\item[RQ5:] How do alternative deep learning mechanisms used in machine translation compare to DeepSTL's transformer architecture?
\end{description}

The last three research questions are relevant to \textbf{Challenge 2}. They are addressed in Section~\ref{sec:machine_translation} and discussed in Section~\ref{sec:discussion}. We employ a corresponding transformer-based NLP architecture, whose attention mechanism enables the efficient training of accurate translators from English to STL. We also compare DeepSTL with other machine translation techniques with respect to translating performance, on synthetic STL training and test data set, and their extrapolations. 

The collected data and the implementation codes in this paper can be found in the provided links\footnote{\label{repo}\emph{Artifact DOI:} \url{10.6084/m9.figshare.19091282} \\ \emph{Code repository:} \url{https://github.com/JieHE-2020/DeepSTL}}.


%% file: 2_related_work.tex
\section{Related Work}
\label{sec:realated}
\noindent
\textbf{From Natural Language to Temporal Logics}\quad
Despite many approaches proposed in the literature~\cite{NelkenF96,DwyerAC99,Ranta11,Kress-GazitFP08,YanCC15,AutiliGLPT15,BrunelloMR19,Nikora2009,LignosRFMK15,Rongjie2015,GhoshELLSS16,FantechiGRCVM94,Dzifcak2009,Christopher2015}, the problem of translating natural language requirements  into temporal logics remains still open~\cite{BrunelloMR19}. The main challenge arises from translating an ambiguous and context-sensitive language into a more precise and context-free formal language. To facilitate and guide the translation process, most of the available methods require the use of predefined specification patterns~\cite{DwyerAC99,Sasha2005,AutiliGLPT15} or of a restricted and more controlled natural language~\cite{SantosCS18,Rongjie2015,Kress-GazitFP08}.
Handling the direct translation of unconstrained natural languages is instead more cumbersome. Other  works~\cite{NelkenF96,Dzifcak2009,LignosRFMK15,GhoshELLSS16} address this problem by translating the  natural language expression first into an intermediate representation. Then, 
the translation process continues by applying a set of manually predefined rules/macros mapping the intermediate representation into temporal logic expressions.  These 
approaches are centered on Linear Temporal Logic (LTL)~\cite{ltl}, a temporal logic suitable to reason about events occurring
 over logical-time Boolean signals.
 
 In this paper we consider instead (for the first time to the best of our knowledge) the problem of automatic translation of unconstrained English sentences into Signal Temporal Logic (STL)~\cite{maler2013monitoring}, a temporal logic that extends LTL with operators to express temporal properties over dense-time real-valued signals.  STL is a well-established formal language employed in both academia and advanced industrial research labs to specify requirements for CPS engineering~\cite{boufaied2021signal,HoxhaAF14}. 

 \noindent
 \textbf{Semantic Parsing}\quad Our problem can be considered an example of semantic parsing~\cite{BrunelloMR19}. This task consists in automatically translating sentences written in a context-sensitive natural language into machine-understandable expressions such as executable code or logical representations. 
 Semantic parsers are automatically learned from a set of utterances written in natural languages that are annotated with the semantic interpretation in the target language. Some relevant toolkits~\cite{BrunelloMR19} available for developing semantic parsers are WASP~\cite{WongM06}, SEMPRE~\cite{sempre}, KRISP~\cite{KateM06}, SippyCup~\cite{sippycup},  and Cornell Semantic Parsing~\cite{Artzi:16spf}.
 Applications of semantic parsing include the translation of questions or commands expressed in  natural-language into Structured-Query-Language (SQL) queries~\cite{Yaghmazadeh0DD17,Fei2014,tang2000,ZelleM96,abs-1709-00103},  Python code~\cite{OdaFNHSTN15},
bash commands~\cite{LinWZE18}, and other domain specific languages~\cite{Quirk2015}.
The main difficulty for this task is to learn the set of semantic rules that can cover all the potential ambiguity arising when translating from a context-sensitive natural language.  Thus, to be effective, this task requires a large training data set. 

In order to cope with the lack of publicly available informal-requirement and formal-specification data sets, we first design a grammar-based generation technique of synthetic data, where each output is a random STL formula and its associated set of possible English translations. Then we address a neural-translation problem, where a deep neural network is trained to predict, given the utterance in English, the optimal STL formula expressing it. Our work leverages  general-purpose deep learning frameworks such as PyTorch~\cite{Ketkar2017} or Tensorflow~\cite{AbadiBCCDDDGIIK16}, and of state-of-the-art solutions based on transformers and their attention mechanisms~\cite{VaswaniSPUJGKP17}.

%% file: 3_STL.tex
\section{Signal Temporal Logic (STL)}
\label{sec:signal}

Signal Temporal Logic (STL) with both {\em past} and {\em future} operators is a formal specification 
formalism used by the academic researchers and practitioners to formalize temporal requirements 
of CPS behaviors. STL allows to express real time requirements of continuous-time 
real-valued behaviors. An example is a simple bounded stabilization property formulated as follows: 
\emph{Whenever V\_In is above 5, then there must exist a time point in the next 10 time units, at which the value of signal V\_Out should be less than 2.}


The syntax of an STL formula $\varphi$ over a set $X$ of real-valued variables is defined by the grammar:
$$
\begin{array}{lcl}
\varphi & := & x \sim u~|~\neg \varphi~|~\varphi_1 \vee \varphi_2~|~\varphi_1 \until_I \varphi_2~|~\varphi_1 \since_I \varphi_2 \\
\end{array}
$$
\noindent where $x \in X$, $\sim \in \{\geq, >, =, <, \leq \}$, $u \in \mathbb{Q}$, 
$I \subseteq [0, \infty)$ is a non-empty interval. For intervals of the form $[a,a]$, we will use the notation $\{a\}$ instead. 
With respect to a signal $w\,{:}\,X\,{\times}\,[0,d)\,{\to}\,\mathbb{R}$, the semantics of an STL formula is described via the satisfiability relation $(w,i) \models \varphi$, indicating that the signal $w$ satisfies $\varphi$ at the time index $i$:
$$
\begin{array}{lcl}

(w,i) \models x \sim u & \leftrightarrow & w(x,i) \sim u \\

(w,i) \models \neg \varphi & \leftrightarrow & (w,i) \not \models \varphi \\

(w,i) \models \varphi_1 \vee \varphi_2 & \leftrightarrow & (w,i) \models \varphi_1 \; \textrm{or} \; (w,i) \models \varphi_2 \\

(w,i) \models \varphi_1 \until_I \varphi_2 & \leftrightarrow & \exists j \in (i + I) \cap \mathbb{T} \; \textrm{:} \; (w,j) \models \varphi_2 \; \\
& & \textrm{and } \forall i < k < j, (w,k) \models \varphi_1 \\

(w,i) \models \varphi_1 \since_I \varphi_2 & \leftrightarrow & \exists j \in (i - I) \cap \mathbb{T} \; \textrm{:} \; (w,j) \models \varphi_2 \; \\
& & \textrm{and } \forall j < k < i, (w,k) \models \varphi_1 \\
\end{array}
$$

We use $\since$ and $\until$ as syntactic sugar for the {\em untimed} variants of the {\em since} $\,\since_{(0,\infty)}$ and 
{\em until} $\,\until_{(0,\infty)}$ operators. From the basic definition of STL, we can derive the following standard operators.
$$
\begin{array}{llcl}
\text{tautology} & \true & = & p \vee \neg p \\
\text{contradiction} & \false & = & \neg \true \\ 
\text{disjunction} & \varphi_1 \wedge \varphi_2 & = & \neg(\neg \varphi_1 \vee \neg \varphi_2) \\
\text{implication} & \varphi_1 \to \varphi_2 & = & \neg \varphi_1 \vee \varphi_2 \\
\text{eventually, finally} & \F_I \varphi & = & \true\,\until_I\,\varphi \\ 
\text{always, globally} & \GG_I \varphi & = & \neg \F_I \neg \varphi \\ 
\text{once} & \opP_I \varphi & = & \true\,\since_I\,\varphi \\ 
\text{historically} & \opH_I  \varphi & = & \neg \opP_I \neg \varphi \\
\text{rising edge} & \rise(\varphi) & = & \varphi \wedge \neg \varphi \, \since \, \true \\
\text{falling edge} & \fall(\varphi) & = & \neg \varphi \wedge \varphi \, \since \, \true \\
\end{array}
$$

We can now formalize the rather verbose English description of the above \textit{Bounded response} requirement, with a succinct STL formula as follows:
$
\begin{array}{c}
\GG (\text{V\_In} > 5 \rightarrow \F_{[0,10]} (\text{V\_Out} < 2 )).
\end{array}
$

This formula can be directly used during the verification of a CPS before it was deployed, or to generate a monitor, checking the safety of the CPS, after its deployment.

%% file: 4_STL_distribution.tex
\section{Empirical STL Statistics}
\label{sec:empirical_stl}

In order to address the relative lack of publicly 
available STL specifications, we develop a synthetic-training-data generator, as described in 
Section~\ref{sec:corpus_construction}. Instead of 
exploring completely random STL sentences, the 
generator should focus on the creation of commonly used
STL specifications. In addition, every STL formula 
shall be associated to a set of natural 
language formulations, with commonly used sentence structure 
and vocabulary.

We analyzed over $130$ STL specifications 
and their associated English-language formulation, from 
scientific papers and industrial documents. 
The investigated literature covers multiple application
domains: 
specification patterns~\cite{boufaied2021signal},
automatic 
driving~\cite{HoxhaAF14,jin2014powertrain,gladisch2019experience}, 
robotics~\cite{liu2017distributed,kapoor2020model,aksaray2016q,liao2020survey,liu2021model}, 
time-series analysis~\cite{chen2020temporal} and 
electronics~\cite{maler2013monitoring,BTS5016-2EKA}. Although 
this literature contains data that is not 
statistically exhaustive, it still provides valuable information 
to guide the design of the data generator and address 
the research question \textbf{RQ1}.

We present our results on the statistical analysis of the STL 
specifications in Section~\ref{subsec:analysis_stl} and 
of their associated natural-language requirements in 
Section~\ref{subsec:analysis_natural}.




\subsection{Analysis of STL Specifications}
\label{subsec:analysis_stl}

We conducted two main types of analysis for the STL specifications 
encountered in the literature: (1)~Identification of 
common temporal-logic templates, and (2)~Computation of the frequency of 
individual operators. During analysis, we made several other relevant observations that we report at the end of this subsection.

\subsubsection{STL-Templates Distribution}
\label{subsubsec:stl_templates}
We identified four common STL templates that we 
call: \textit{Invariance/Reachability}, 
\textit{Immediate response}, 
\textit{Temporal response} and \textit{Stabilization/Recurrence}.


\vspace*{1mm}\noindent \textbf{Invariance/Reachability template:} 
Bounded and unbounded invariance and reachability 
are the simplest temporal STL properties. They have the form $\GG \varphi$, $\GG_{[a,b]} \varphi$,
$\F \varphi$ or $\F_{[a,b]} \varphi$,
where $\varphi$ is an atomic predicate. We provide one 
example of bounded-invariance (BI)~\cite{jin2014powertrain}, 
and one example of 
unbounded-reachability (UR)~\cite{kapoor2020model} 
specification, respectively, as encountered 
in our investigation:
$$
BI: \GG_{[\tau_{s},T]} (\mu < c_{l}), \quad
UR: \F (x > 0.4)
$$

\noindent \textbf{Immediate response template:} This template represents 
formulas of the form $\GG (\varphi \rightarrow \psi)$, 
where $\varphi$ and $\psi$ are atomic propositions or their Boolean combinations. Except for the starting $\GG$ operator, there are no other temporal operators in the formula. An example of 
an immediate response (IR) specification is the one
from~\cite{boufaied2021signal}: 
$$
IR: \GG (\text{not\_Eclipse} = 0 \rightarrow \text{sun\_currents} = 0).
$$

\noindent \textbf{Temporal response template:} This template represents 
formulas of the form $\GG (\varphi \rightarrow \psi)$, 
where $\varphi$ and $\psi$ can have non-nested temporal 
operators. We illustrate several TR specifications that 
we encountered in the literature.  They all belong to this class:
$$
\begin{array}{lll}
TR1: \GG (\rise (\text{Op\_Cmd} = \text{Passive}) \rightarrow \F_{[0,500]} \text{Spd\_Act} = 0)  \\
TR2: \GG (\text{currentADCSMode} = \text{SM} \rightarrow P \  \until_{[0,10799]} \ \neg P ) & \text{\cite{boufaied2021signal}} \\
TR3: \GG (\rise (\text{gear\_id} = 1) \rightarrow \GG_{[0,2.5]} \neg \fall (\text{gear\_id} = 1)) & \text{\cite{HoxhaAF14}} \\
\end{array}
$$
\noindent In $TR2$ above, $P \equiv \text{real\_Omega} - \text{target\_Omega} = 0$.

\vspace*{1mm}
\noindent \textbf{Stabilization/Recurrence template:} These templates represent 
formulas allowing one nesting of the temporal operators. 
Typical nesting is $\GG\F\,\varphi$ for recurrence (RE), and $\F\GG\,\varphi$ for stabilization (ST),
with their bounded counterparts. Here $\varphi$ is a non-temporal 
formula. The following specifications from the literature are in this category:
$$
\begin{array}{ll}
ST: \F_{[0,14400]} \GG_{[4590,9963]} \ (x_{10} \geq 0.325 ) & \text{\cite{chen2020temporal}} \\
RE: \GG_{[0,12]} ( \F_{[0,2]} \ \text{regionA} \ \wedge \ \F_{[0,2]} \ \text{regionB}) & \text{\cite{liao2020survey}} \\
\end{array}
$$

\noindent \textbf{Other formulas:} These
are formulas that do not fall into any of the above categories.
The following specification belongs to this class:
$$
\GG( \rise (\varphi_1) \rightarrow \F_{[0,t_{1}]} (\rise (\varphi_2) \ \wedge (\varphi_2 \ \until_{[t_{2},t_{3}]} \ \varphi_3 )) )
$$
It captures the following requirement: 
\emph{Whenever the precondition $\varphi_{1}$ becomes true, 
there is a time within $t_1$ units where $\varphi_2$ 
becomes true and continuously holds until $\varphi_3$ 
becomes true within another interval $[t_2, t_3]$.}
This pattern is used in the electronics field~\cite{maler2013monitoring} to describe the situation where one digital signal tracks another~\cite{BTS5016-2EKA}.

\vspace*{1mm}
\noindent \textbf{Statistics:} 
We encountered $39$ \textit{Invariance/Reachability} ($30.0\%$), 
$27$ \textit{Immediate response} ($20.8\%$), $33$ \textit{Temporal 
response} ($25.4\%$) and $31$ \textit{Stabilization/Recurrence} ($23.8\%$) 
templates. The category of Other templates is orthogonal to the first four ones since it includes ad hoc formulas. There are overall 13 ($39.4\%$) \textit{Temporal 
response} and 6 ($19.3\%$) \textit{Stabilization/Recurrence} templates belonging to this type.



\subsubsection{STL-Operators Distribution}
\label{subsubsec:stl_operator_distribution}
\begin{figure}[H]
\centering
\begin{tikzpicture}
{\small
\tkzKiviatDiagram[scale=0.3,label distance=.4cm,
        radial  = 1,
        gap     = 1.6,  
        lattice = 5]{$\ \ \ x = u$,$x > u$, $x \geq u$,
        $x < u$, $x \leq u$, $\rise(\varphi)$,$\fall(\varphi)$,
        $\F_I \varphi$, $\GG_I \varphi$, $\varphi_1 \until_I \varphi_2$,
        $\opP_I \varphi$, $\opH_I \varphi$, $\varphi_1 \since_I \varphi_2$, 
        $\neg \varphi $, $\varphi_1 \wedge \varphi_2$, $\varphi_1 \vee \varphi_2$,  
        $\varphi_1 \rightarrow \varphi_2$
          }
\tkzKiviatLine[thick,mark=ball,mark size=4pt,color=darkgray,
              fill=green!25,opacity=.5](4.84/1.2, 1.4/1.2, 2.12/1.2, 1.84/1.2, 2.12/1.2, 1.96/1.2, 0.88/1.2, 2.72/1.2, 6.08/1.2, 1.04/1.2, 0.2/1.2, 0.44/1.2, 0.32/1.2, 1.12/1.2, 4.2/1.2, 0.52/1.2, 3.52/1.2)   
\tkzKiviatGrad[prefix=,unity=30](1)}
\end{tikzpicture}
\caption{Frequency Distribution of STL Operators.}
\label{fig:operator_occurrence_times}
\vspace{-1ex}
\end{figure}
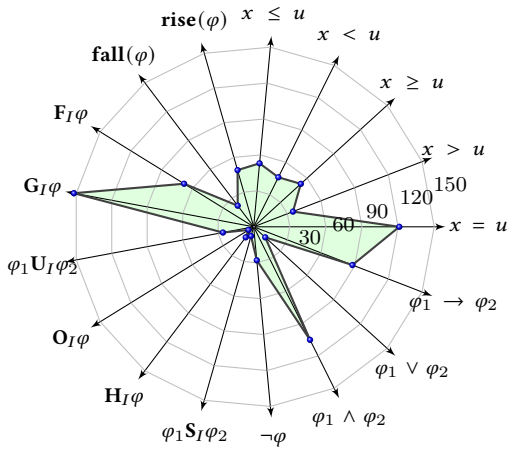

We investigated the distribution of the STL-operators as
encountered in the specifications found in the above-mentioned literature.
Figure~\ref{fig:operator_occurrence_times} summarizes 
our results.



We first discuss the (atomic) numeric predicates.
We observe that the equality operator occurs 
more often in numerical predicates 
then the other ($\leq$, $<$, $\geq$, $>$) relations. 
This happens because many specifications refer 
to discrete mode signals and equality is used to 
check if a discrete variable is in a given mode.


Conjunction and implication are the two most 
frequently used Boolean operators. Conjunction 
is often used to specify that a signal must lie within 
a given range. Implication is used in a pretty wide range of 
response specifications.


The $\rise$ and $\fall$ operators 
are typically used in front of an atomic predicate
(for example $\rise(x > \mu)$). The frequency of 
rising edges is higher than that of falling edges, 
which can be explained by the fact that many specifications
refer to time instants where a condition starts holding, 
rather than when it stops holding.


The $\GG$ operator has a much higher frequency than any other temporal operator.
This is not surprising because $86.2\%$ (112/130) of the
specifications are invariance or response or recurrence properties 
that start with an always operator. The $\F$ operator 
ranks second and is often used in specifications 
of robotic applications to define reachability 
objectives. We also remark that ``eventually'' is used 
in bounded/unbounded and stabilization properties.

We finally observe that future temporal 
operators ($\GG$, $\F$, $\until$) are used more often than their past counterparts ($\opP$, $\opH$, $\since$)
and that unary temporal operators 
($\GG$, $\F$, $\opH$, $\opP$) are used more often 
than the binary ones ($\until$, $\since$). 
These two observations are explained by the fact that 
most declarative specifications have a natural 
future flavor (a trigger now implies an obligation that 
must be fulfilled later) and unary temporal operators 
are easier to understand and handle.




\subsubsection{Other Observations}
\label{subsubsec:other_observations}
In this section, we discuss additional findings that 
we discovered during the analysis of the STL specifications occurring
in the literature:
\begin{itemize}
    \item We found a frequent usage of the pattern 
    $|x -y|$ to denote the pointwise distance between 
    signals $x$ and $y$, especially in the motion control
    applications~\cite{gladisch2019experience,liu2017distributed,kapoor2020model}.
    \item Some publications use abstract predicates to denote 
    complex temporal patterns, without providing their detailed
    formalization. One such example is the use of the predicate 
    $\textsf{spike}(x)$ to denote a spike occurring within the 
    signal $x$~\cite{boufaied2021signal}.
    \item It is relatively common in the literature to decompose a complex STL 
    specification into multiple simpler ones, by giving 
    a name to a sub-formula and using that name as an 
    atomic proposition in the main formula.
    \item Time bounds in temporal operators and signal 
    thresholds are sometimes given as parameters, rather 
    than constants.
    \item $\rise$, $\fall$ and past temporal operators are normally used as pre-conditions, while future operators are often used as post-conditions. Negation is used conservatively, e.g., $\neg\fall$ is used to represent a particular stabilized condition should hold for a designated time interval~\cite{maler2013monitoring}.
\end{itemize}

\subsection{Analysis of NL Specifications}
\label{subsec:analysis_natural}

In this section we investigate the usage of natural 
language (NL) in the literature to express informal 
requirements, which are then formalized using STL. 
In particular, we identified the English vocabulary used 
to formulate STL operators and sentences, and 
studied the quality, 
accuracy and preciseness of the language. 

\subsubsection{English formulation of STL sentences} 
\label{subsubsec:english_formulation}
We considered several aspects (e.g.,~nouns, verbs, adverbs, etc.) when studying the use 
of natural language in the formulation of:
\begin{itemize}
    \item Numeric (atomic) predicates,
    \item Temporal operators (phrases),
    \item Specific scenarios (e.g.,~a rising/falling edge).
\end{itemize}


The main outcome of this analysis is that the language 
features used in the studied requirements are unbalanced 
and sparse 
and that it is hard to identify a general recurring pattern.
We illustrate this observation with two representative 
examples:
%
\begin{itemize}
    
    
    \item \textbf{Example 1}: We counted different English 
    utterances to express the semantics of $x > \mu$. 
    The most frequently used collocation is 
    ``be above'', which appears overall four times. 
    Next comes ``increase above'' (somewhat ambiguous because this may also represent a rising edge), which is used two times. 
    Then ``be higher than'', ``be larger than'' and ``be greater than'' are only used 
    once respectively. However, we do not find any 
    requirements using other synonymous expressions 
    like ``be more than'' or ``be over''.
    
    \item \textbf{Example 2}: We observed that two 
    temporal adverbs are frequently used to express 
    $\GG_{[0,t]}$ and $\opH_{[0,t]}$, which are 
    ``for at least $t$ time units ($s, ms$, etc.)'' 
    (eight times) and ``for more than $t$ time units'' 
    (six times). However, other reasonable 
    possibilities like ``for the following/past $t$ time units'' are not found.
\end{itemize}

The sparsity and lack of balance 
may be a consequence of the relative small base of publicly available 
literature that defines this type of requirements.  
Despite the fact that the findings of this analysis may not 
be sufficiently representative, we can still use the 
outcomes to improve our synthetic generation of examples.



\subsubsection{Language Quality}
For the English requirements found in the literature, of 
particular interest is the language quality: 
How accurately does a requirement reflect 
the semantics of its corresponding STL formula? 
Given this criterion, we classify the studied English requirements into {\em Clear}, {\em Indirect} and 
{\em Ambiguous} requirements.

\vspace*{1mm}
\noindent \textbf{Clear:} These requirements have 
a straightforward STL formalisation that results in an unambiguous specification without room for interpretation.
An example of a clear requirement is the sentence: \emph{If the value of signal \textit{control\_error} is less than $10^{\circ}$, then the value of signal  \textit{currentADCSMode} shall be equal to  \textit{NMF}}~\cite{boufaied2021signal}. The resulting 
STL specification is given by the formula:
$$
\GG (\text{control\_error} < 10 \rightarrow \text{currentADCSMode} = \text{NMF})
$$

\noindent \textbf{Indirect:} These requirements need an expert to translate them into an STL formula that 
faithfully captures the intended meaning. They 
typically assume some implicit knowledge that must 
be added to the formal specification from the context. 
An example is the sentence: \emph{The vehicle shall stay within the lane boundaries, if this is possible with the actuators it is equipped with}~\cite{gladisch2019experience}. This is an indirect 
requirement formalized using the following 
STL formula:
$$
\GG (\tau < \tau_{max} \rightarrow \textbf{P})
$$
\noindent Here $\textbf{P}$ is the contextual sub-formula: $\textit{vehicle} \subseteq \textit{corridor}$.

\vspace*{1mm}
\noindent \textbf{Ambiguous:} These requirements lack 
key information that cannot be easily inferred from 
the context and that must be extracted from external 
sources, such as tables, figures, timing diagrams, or experts. 
They use vague and ambiguous language, 
and can have multiple interpretations. An example is the following sentence: \emph{To prevent the destruction of the device by avalanche due to high voltages, there is a voltage clamp mechanism $Z_{DS(AZ)}$ implemented, that limits negative output voltage to a certain level $V_S-V_{DS(AZ)}$. Please refer to Figure 10 and Figure 11 for details}~\cite{BTS5016-2EKA}. This is an ambiguous requirement that can be translated to the following STL formulas:
$$
\begin{array}{c}
\GG (V_{OUT} < V_{GND} \wedge I_L > 0 \rightarrow V_{OUT} = V_S-V_{DS(AZ)})\\
\GG (V_{OUT} < V_{GND} \rightarrow V_{OUT} = V_S-V_{DS(AZ)})\\

\end{array}
$$
\noindent The English requirement only vaguely mentions the post-condition. The pre-condition characterizes the drop of voltage $V_{OUT}$ below $V_{GND}$ when the inductive load is being switched off. This is obtained from the previous context and Figure 11 of ~\cite{BTS5016-2EKA} with some physical knowledge that inductive current has to change smoothly.

We encountered $46$ \textit{Clear} ($35.4\%$), 
$43$ \textit{Indirect} ($33.1\%$), and $41$ \textit{Ambiguous} ($31.5\%$) 
English requirements.

%% file: 5_corpus_construction.tex
\section{Corpus Construction}
\label{sec:corpus_construction}
This section addresses research question $\textbf{RQ2}$. It first introduces a new method for the
automatic generation of STL sentences and their 
associated natural language requirements. The 
generator incorporates the outcomes from Section~\ref{sec:empirical_stl} for improved results. 
Finally, we use this method to do the actual generation 
of STL-specification/NL-requirements pairs. 

\subsection{Corpus Generation}
\label{subsec:stl_eng_generation}
In the following, we propose an automatic procedure for randomly generating synthetic examples. Each example consists of: (1)~An STL formula, 
and (2)~A set of associated sentences in English 
that describe this formula. We associate multiple 
natural-language sentences to each formal STL 
requirement to reflect the fact that formal specifications
admit multiple natural language formulations.
%
We illustrate this observation using the 
\textit{Bounded response} specification from Section~\ref{sec:signal}, formalized as the STL formula below:
\vspace*{-1mm}
$$
\GG(\text{V\_In} > 5 \rightarrow \F_{[0,10]} ( \text{V\_Out} < 2 ))
$$
%
%
%
This admits multiple synonymous English 
formulations, including:
	

\begin{itemize}

\item Globally, if the value of V\_In is greater than 5, then finally the value of V\_Out should be smaller than 2 at a time point within 10 time units.

\item It is always the case that when the signal V\_In is larger than 5, then eventually at sometime during the following 10 time units the signal V\_Out shall be smaller than 2.
\end{itemize}

This example shows that two NL formulations of the same STL formula can be very different, making the generation of synthetic 
examples a challenging task. The systematic translation 
of unrestricted STL is indeed extremely difficult, especially 
for specifications that include multiple nesting of 
temporal operators. In practice, deep nesting of temporal 
formulas is rarely used because the resulting specifications 
tend to be difficult to understand.

Hence, we first restrict STL to a rich but well-structured sub-fragment that facilitates a fully automated translation, 
while at the same time covering commonly used specifications. 

\subsubsection{Restricted-STL Fragment}
\label{subsubsec:support_grammar}

In this subsection, we present the restricted fragment of STL that 
we support in our synthetic example generator.  
%
We define this fragment using three layers that 
can be mapped to the syntax hierarchy identified in Section~\ref{subsubsec:stl_templates}. 

The bottom layer, called 
\textbf{simple-phrase} (\textbf{SP}) 
layer, consists of: (1)~\textbf{Atomic propositions} 
($\bm{\alpha}$) including rising and falling edges 
and (2)~\textbf{Boolean combinations} of up to two 
atomic propositions. 
\begin{align*}
\alpha \ \ := \ \
& \ x \circ u~|~\neg (x \circ u)~|~\rise(x \circ u)~|~\fall(x \circ u)~| \\ 
& \ ~\neg \rise(x \circ u)~|~\neg \fall(x \circ u) \\ 
\SP \ \ := \ \
& \ \alpha~|~\alpha \wedge \alpha~|~\alpha \vee \alpha~
\end{align*} 
\noindent where $x$ is a signal name, $u$ is a constant or a mode name, and 
$\circ \in \{ <, \le, =, \ge, > \}$.

The middle layer, which we call 
\textbf{temporal-phrase} (\textbf{TP}) layer, 
admits the specification of temporal formulas 
over simple phrases:
\begin{align*}
\TP \ \ := \ \
& \ {\TP}^{\prime}~|~\neg {\TP}^{\prime}~|~\rise~{\TP}^{\prime}~|~\fall~{\TP}^{\prime}~|~\neg \rise~{\TP}^{\prime}~|~\neg \fall~{\TP}^{\prime}\\ 
{\TP}^{\prime} \ \ := \ \
& \ \UTO_I~(\alpha)~|~(\alpha)~\BTO_I~(\alpha)
\end{align*} 
where $\UTO \in \{\F, \GG, \opP, \opH\}$ and $\BTO \in \{\until, \since\}$ are unary and binary temporal operators, respectively. $I$ is an interval of the form $[t_1,t_2]$ with 
$0 \le t_1 < t_2 \le \infty$. This can be omitted if $t_{1}=0$ and $t_{2}= \infty$.

The top layer, which we call single \textbf{nested-temporal-phrase (NTP)} layer, allows the formulation of formulas with a single nesting of 
a subset of temporal operators:
\begin{align*}
\NTP \ \ := \ \
& \ \F_I\GG_I(\alpha)~|~\GG_I\F_I(\alpha)
\end{align*} 
where $I$ follows the same definition as mentioned above.


Finally, with an auxiliary syntactical component $\PP \ := \ \SP~|~\TP$, formula $\psi$ defines the \textbf{supported fragment} of STL that we map to the four template categories discussed in Section~\ref{subsubsec:stl_templates}. 
%
\begin{align*}
\psi \ \ := \
& \ \ \ \GG_I(\SP)~|~\F_I(\SP) \ \ \ \   (Invariance/Reachability) \\
& \ |~\GG~(\SP \to \SP)  \quad \quad   (Immediate \ response) \\
& \ |~\GG~(\PP \to \TP) \quad \quad \ \,  (Temporal \ response) \\
& \ |~\GG~(\PP \to \NTP) \ \ \, \quad    (Stabilization/Recurrence)
\end{align*} 

This fragment balances between \textit{generality}, needed to express common-practice requirements, and \textit{simplicity}, needed 
to facilitate the automated generation of synthetic examples. It results in the following restrictions: 
(1)~We allow the conjunction and disjunction of only two atomic propositions, (2)~Only one atomic proposition is allowed 
inside a temporal operator in \textbf{TP}, (3)~We do not allow 
Boolean combinations of \textbf{SP} and \textbf{TP} formulas, 
and (4)~Formulas outside the four mentioned templates are not supported.

\begin{figure}[ht]
	\includegraphics[width=0.4\textwidth]{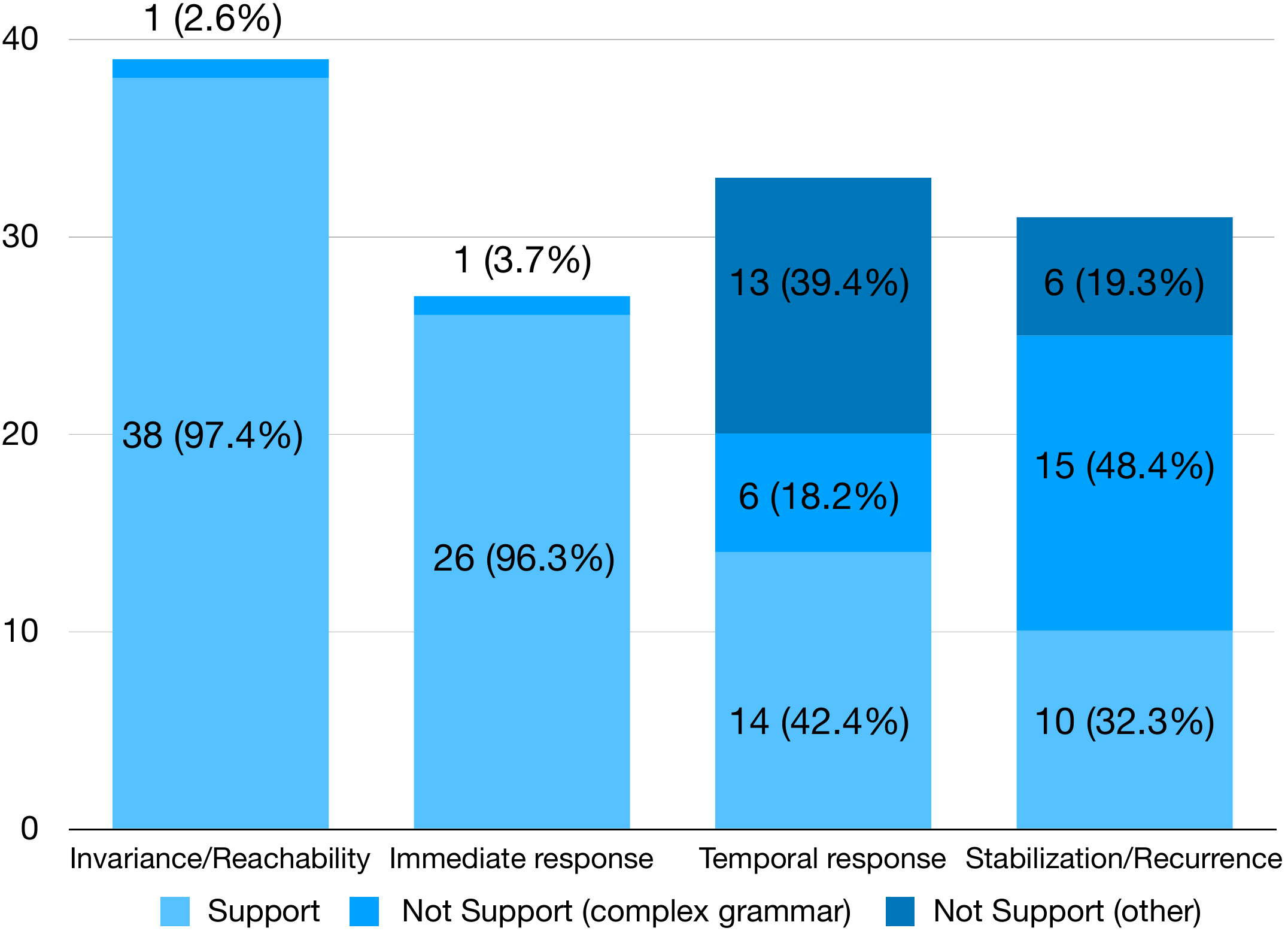}
	\caption{STL Template Support Summary.}
	\label{fig:STL_template_support_summary}
\end{figure}

By relating the generator-fragment $\psi$ to the empirical statistics in Section~\ref{subsubsec:stl_templates}, Figure~\ref{fig:STL_template_support_summary} summarizes for each syntactical category, the proportion of templates that the fragment can support.

The generator nearly supports all \textit{Invariance/Reachability} templates appearing in our database. For \textit{Immediate response} ones, there is one template missing due to restriction (1). For \textit{Temporal response} templates, we are able to support $42.4\%$ of them. For the not supported ones, $18.2\%$ use a complex grammar that violates restriction (2) and (3), while the remaining ones ($39.4\%$) belong to the \emph{other} category for ad hoc purposes.  Concerning nesting formulas, we only consider \textit{Stabilization/Recurrence} templates. Other combinations such as $\F \F \varphi $ or $\F \varphi_1 \until \varphi_2 $ are not supported: $48.4\%$ of them are in the \emph{complex grammar} group, while the left $19.3\%$ are in the \emph{other} category.

\subsubsection{Random-Sampling STL Formulas}
This short subsection briefly describes how we sample STL specifications from the restricted 
fragment. The main idea is to decorate the grammar rules with probabilities according to the template distribution collected in Section~\ref{subsec:analysis_stl} and the operator distribution shown in Figure ~\ref{fig:operator_occurrence_times}.

Consequently, we use the probabilities described in Section~\ref{subsec:analysis_stl} to generate the four categories of fragment $\psi$, which will naturally make the $\GG$ operator rank first to a large extent, followed by the $\F$ operator, regarding to usage frequency. The frequencies of the other operators within these categories are as discussed above.

\subsubsection{Translating STL into English}

The main translation strategy linked to~\ref{subsec:analysis_natural} is as follows. For the predicates used to express logical relations in the bottom layer, we use (with some reservations mainly with regards to accuracy) the frequencies of Section~\ref{subsubsec:english_formulation}. This way, the translation candidates are selected with different weights. For the others, such as the adverbs specifying temporal information, we incorporated relevant English utterances encountered in our database, on the condition that the generation and recognition can be done with both accuracy and fluency. Furthermore, we reserved enough space 
to add synonymous utterances that can be typically used but that are not included in the database.

In order to systematically organize the translation and maximize language flexibility, we start with the translation of atomic propositions (defined as $\alpha$ in~\ref{subsubsec:support_grammar}) in the bottom syntactical layer, and use this as a pivot to tackle temporal phrases and their nesting scenarios in the middle and top layers.

\begin{figure}[ht]
	\includegraphics[width=0.482\textwidth]{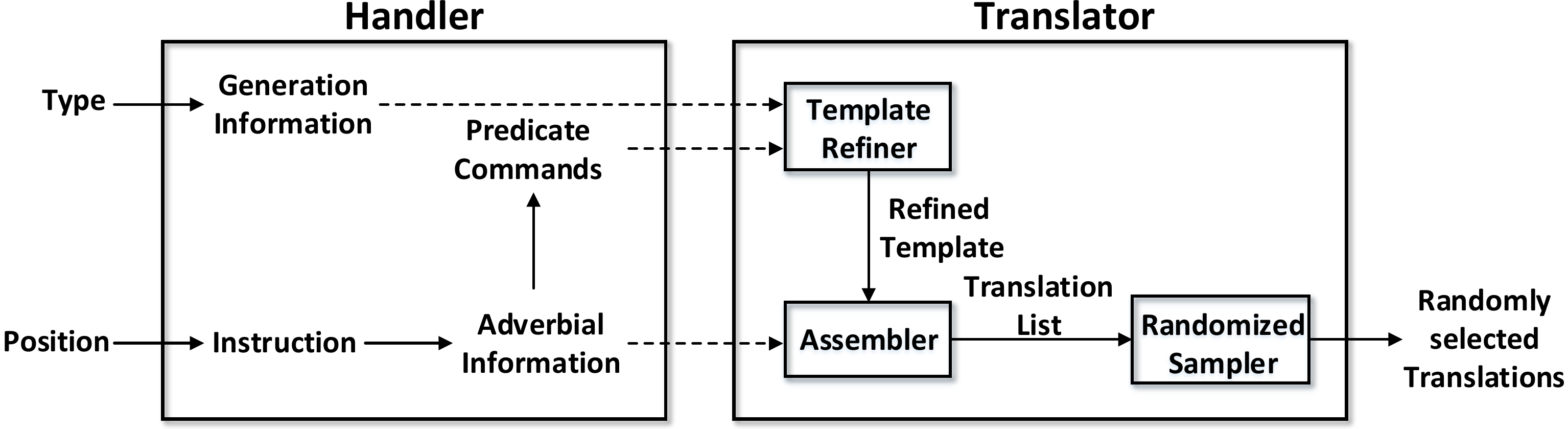}
	\caption{Translation Procedure for Atomic Propositions.}
	\label{fig:translation_procedure}
\end{figure}

\vspace*{1mm}
\noindent
\textbf{Bottom layer.}\quad
The English counterparts of atomic propositions typically consist of a subject, a predicate, and an object. They are indispensable in each English sentence. Hence, their variations especially in the predicate (including the choice of verbs, formats, tenses, and their active/passive voice) are considered first. The workflow for the organisation of their translations, which is divided into a \textit{Handler} and a \textit{Translator}, is illustrated in Figure~\ref{fig:translation_procedure}.

The Handler, as a preprocessor, takes the \textit{Type} and \textit{Position} information as inputs. \textit{Type} is a branch of $\alpha$ used to compute and output the \textit{Generation Information}. This includes an \textit{index} (it triggers a corresponding translation strategy), \textit{identifiers}, \textit{numbers}, and the \textit{STL expression} of a randomly generated atomic proposition.

\textit{Position} specifies the location of the proposition. This determines whether the translation states a certain scenario - if it is a condition (before implication symbol ``$\to$''), or if it emphasizes that a property has to hold with a satisfied condition (after ``$\to$''). In the latter case, modal verbs like ``should'' or``must'' will be used, and often together with adverbial modifiers like ``instantly'' or ``without any delay'' in case of \textit{Immediate response} formulas. This information is embedded into \textit{Predicate Commands}, incorporating the choice of verbs, their format, and the use of modal verbs and of adverbial modifiers. 

\begin{figure}[ht]
	\includegraphics[width=0.485\textwidth]{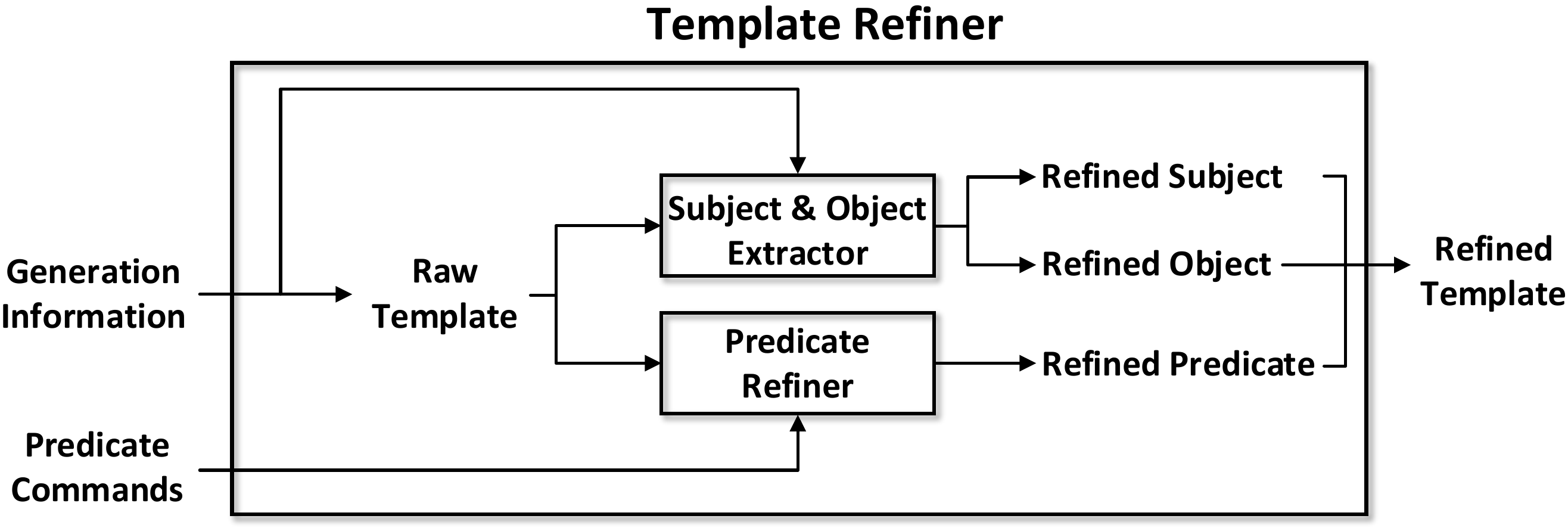}
	\caption{Template Refiner.}
	\label{fig:template_refiner}
\end{figure}

\textit{Generation Information} and \textit{Predicate Commands} are sent to the \textit{Template Refiner} (inside Translator), whose architecture is shown in Figure~\ref{fig:template_refiner}. Here, the subject and object placeholders within the templates can be replaced by randomly generated identifiers and numbers. The verbs associated to predicates are changed to their proper format, and are decorated with adverbs when applicable. 

In the next step, the \textit{Assembler} module of the Translator completes the refined templates into a complete sentence that also includes adverbial modifiers. Finally, the \textit{Randomized Sampler} module of the Translator, samples a designated number of sentences from the overall translation list.

\vspace*{1mm}
\noindent
\textbf{Middle/Top layer.}\quad
The translation approach presented above is extended to temporal phrases in a straight-forward manner, because the sentences generated by the bottom layer, can be reused except for the need to add adverbial modifiers, and enrich the verb tenses according to the temporal operators.

\begin{figure}[ht]
	\includegraphics[width=0.48\textwidth]{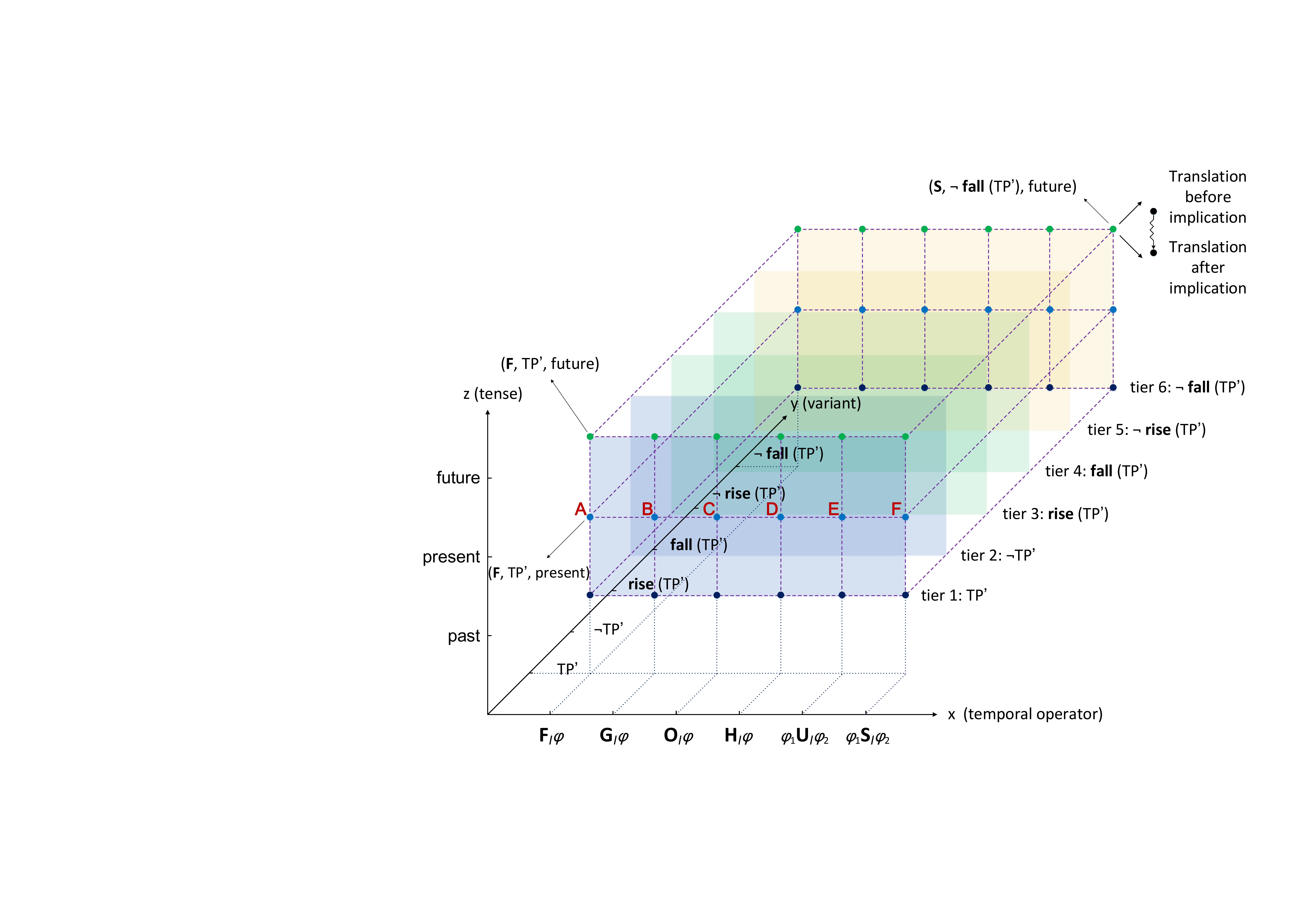}
	\caption{Translation of Temporal Phrases.}
	\label{fig:translate_tp}
\end{figure}

The temporal aspects nevertheless do increase the 
translation complexity. We need to consider three 
orthogonal aspects (dimensions) as shown in Figure~\ref{fig:translate_tp}. The $x$-axis represents 
the six STL temporal operators from the \textbf{TP} 
layer, the $y$-axis their variants preceded by the negation, 
rising or falling edge operators, and the $z$-axis 
the choice of a verb tense in English for specific 
temporal operators. Hence, a node in Figure~\ref{fig:translate_tp} represents a specific 
combination of these three aspects. 


We adopt a slicing approach to tackle the complexity. We first process nodes A-F with present tense, where the six temporal operators are used individually. Then we enrich the usage of verb tenses according to the semantics of a particular operator and its nesting situation. This results in tier ${\TP}^{\prime}$ while preserving language flexibility. The same approach is used for processing unary operators in tier $\neg {\TP}^{\prime}$. The semantics of direct negation of binary operators, rising/falling edges and their negations for layer \textbf{TP} are complicated. Considering their relatively low usage frequency, we provide several fixed templates to facilitate their translations.

\subsection{Corpus Statistics}
Following the approach described in Section~\ref{subsec:stl_eng_generation},
we have automatically generated a corpus consisting of 120,000 STL-English pairs where each pair consists of a randomly generated STL formula and one of its generated translation in natural language. 

\subsubsection{STL-Formula Statistics}
\begin{figure}[ht]
\centering
\begin{tikzpicture}
{\small
\tkzKiviatDiagram[scale=0.29,label distance=.4cm,
        radial  = 1,
        gap     = 1.6,  
        lattice = 5]{$\ \ \ x = u$,$x > u$, $x \geq u$,
        $x < u$, $x \leq u$, $\rise(\varphi)$,$\fall(\varphi)$,
        $\F_I \varphi$, $\GG_I \varphi$, $\varphi_1 \until_I \varphi_2$,
        $\opP_I \varphi$, $\opH_I \varphi$, $\varphi_1 \since_I \varphi_2$, 
        $\neg \varphi $, $\varphi_1 \wedge \varphi_2$, $\varphi_1 \vee \varphi_2$,  
        $\varphi_1 \rightarrow \varphi_2$
          }
\tkzKiviatLine[thick,mark=ball,mark size=4pt,color=darkgray,
              fill=blue!25,opacity=.5](4.1284, 2.1106, 2.0786, 2.0816, 2.1176, 2.4978, 1.0082, 2.2166, 5.5486, 0.3618, 0.2192, 0.2154, 0.2042, 1.5446, 3.1224, 0.6784, 3.35)   
\tkzKiviatGrad[prefix=,unity=25,suffix=K](1)  }
\end{tikzpicture}
\caption{Frequency of STL operators in the corpus.}
\label{fig:frequency_operator_dataset}
\vspace{-1ex}
\end{figure}
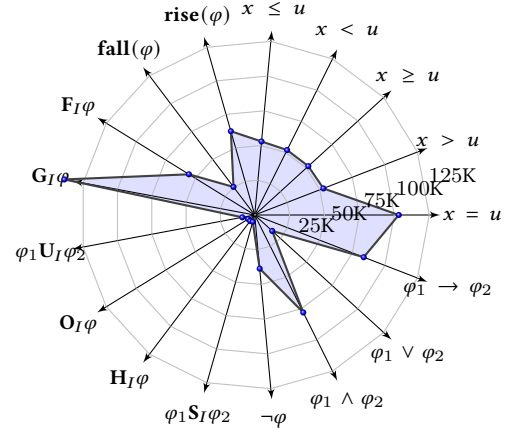

In Figure~\ref{fig:frequency_operator_dataset} we provide the frequencies of the STL operators in our synthetic dataset (above corpus). As one can see they are largely consistent with the ones in Figure~\ref{fig:operator_occurrence_times}. As before, the most frequent STL operator is the global temporal  operator $\GG_I \varphi$ with 138,715 occurrences. The least frequent STL operator is the $\varphi_1 \since_I \varphi_2$ temporal operator with 5,105 occurrences. While this frequency differs a bit from Figure~\ref{fig:operator_occurrence_times}, it is still consistent with the empirical results.

Table~\ref{tbl:stl_statistics} shows the statistics of templates and subformulas in the generated corpus. As mentioned in Section \ref{subsec:analysis_stl}, an STL template is defined as the parse tree of a formula without its leaves. For example, the template for the formula $\varphi~{=}~\GG(\text{In}> 5 \rightarrow \F_{[0,10]}\,\text{Out} < 2 )$ is 
$\GG(\varphi_1 \rightarrow \F_{[0,10]} \varphi_2 )$. Each formula has a finite number of subformulas.  For example the formula $\varphi$ above has five subformulas:   
$\varphi_5~{=}~\GG(\text{In}\,{>}\,5~\rightarrow~\F_{[0,10]}\,\text{Out} < 2 )$,
$\varphi_4~{=}~\text{In}\,{>}\,5~{\rightarrow}~\F_{[0,10]}\,\text{Out}\,{<}\,2$,
$\varphi_3~{=}~\F_{[0,10]} \text{Out}\,{<}\,2$,
$\varphi_2~{=}~\text{Out}\,{<}\,2$, and
$\varphi_1~{=}~\text{In}\,{>}\,5$. 

\begin{table}[ht]
\caption{STL Formula Statistics: \# unique STL formulas, \# unique STL templates, \# subformulas for each formula.}
\vspace{3ex}
\setlength{\belowcaptionskip}{5pt}
\centering
\begin{tabular}{c|c|c|c|c|c}
\hline
\multirow{2}{*}{\begin{tabular}[c]{@{}l@{}}\# formulas\end{tabular}} & \multirow{2}{*}{\begin{tabular}[c]{@{}l@{}}\# templates\end{tabular}} & \multicolumn{4}{c}{\# subformula per formula} \\
\cline{3-6} 
                                                                     &                                                                & min    & max   & avg.   & median \\ \hline
120,000	&  5,852	&   2	&   18	&   6.98  	&   7   \\ \hline
\end{tabular}

\label{tbl:stl_statistics}
\end{table}


Table~\ref{tbl:mapping_statistics} shows the mutual mapping relation between STL operators and STL formulas in our corpus. 
We count for each STL operator, how many formulas it has appeared in.  This produces the containment
statistics shown in the last three columns. 

\begin{table}[ht]
\caption{STL-Formula Mapping Statistics: \# STL operators for each formula, \# STL formulas for
each operator.}
\vspace{3ex}
\setlength{\belowcaptionskip}{5pt}
\centering
\begin{tabular}{c|c|c|c|c|c}
\hline
\multicolumn{3}{l|}{\# STL oper. per formula} & \multicolumn{3}{l}{\# formulas per STL oper.} \\ \hline
avg.           & median           & max           & avg.    & median    & \multicolumn{1}{l}{max}    \\ \hline
6.98           &  7                & 18          & 42,120.3        & 45,125        & 101,870                            \\ \hline
\end{tabular}
\label{tbl:mapping_statistics}
\end{table}


Since identifiers and constants frequently appear in our corpus, we also analyzed their frequency, as shown in Table~\ref{tbl:id_constant_statistics}. 

\begin{table}[ht]
\caption{Identifier and Constants Statistics: average number of identifiers per formula, \# of chars used per identifier, \# number of digits used per constant.}
\vspace{3ex}
\setlength{\belowcaptionskip}{5pt}
\centering
\begin{tabular}{c|c|c|c|c|c|c}
\hline
\multirow{2}{*}{\begin{tabular}[c]{@{}l@{}}\# identifiers\\ per formula \end{tabular}} & \multicolumn{3}{l|}{\# chars per identifier} & \multicolumn{3}{l}{\# digits per constant} \\ 
\cline{2-7}                                                                                              
 																					   &   min          & avg.          & median         & min          & avg.          & median    \\ \hline
2.59	&  1	&   5.50 	&   5	&   1  	&   2.31  	&   2   \\ \hline
\end{tabular}
\label{tbl:id_constant_statistics}
\end{table}


\subsubsection{Natural-Language Statistics}
The statistical results of the natural language in our corpus are shown in Table~\ref{tbl:nl_statistics}. There are only 265 different effective English words (considering word variants, not including signal names which are strings generated randomly), constituting a relatively small vocabulary. This is understandable because most English words are used to express the logical relation in STL, the number of which is thus limited. Besides, Table~\ref{tbl:nl_statistics} records the statistics of effective word numbers in all English sentences. It also counts for each English word, the number of English sentences using it. 

\begin{table}[ht]
\caption{English Statistics: \# unique sentences,  \# unique words, \# words per sentences and \# sentences per word.}
\vspace{3ex}
\setlength{\belowcaptionskip}{5pt}
\centering
\begin{tabular}{c|c|c|c|c|c}
\hline
\multirow{2}{*}{\# sent.} & \multirow{2}{*}{\# word} & \multicolumn{2}{l|}{\# words per sent.} & \multicolumn{2}{l}{\# sent. per word} \\ \cline{3-6} 
                          &                          & avg.              & median              & avg.              & median             \\ \hline
120,000                        & 265                       & 38.49                & 37                  & 14,220.28                & 4,555.5                 \\ \hline
\end{tabular}
\label{tbl:nl_statistics}
\end{table}






%% file: 6_machine_translation.tex
\section{Machine Translation}
\label{sec:machine_translation}

In order to answer questions \textbf{RQ3-5}, we take advantage of the corpus generated as discussed in the previous sections, to develop DeepSTL, a tool and technique for the translation of informal requirements given as free English sentences, into STL.  DeepSTL employs a state-of-the-art transformer-based neural-translation technique, to train an accurate attentional translator. We compare the performance of DeepSTL with other NL translator architectures on the same corpus, and we also investigate how they are able to extrapolate to sentences out of the corpus.


\subsection{Neural Translation Algorithms}
The translation of natural language into STL formulas can be abstracted as the following probabilistic problem. Given an encoding sequence $\textbf{e}= (e_1, e_2, ..., e_m)$ from the source language (English requirements), a decoding sequence $\textbf{s} = (s_1, s_2, ..., s_n)$ from the target language (STL formulas) generates all of its tokens $s_k$ conditioning on the decoded history of the target sequence $s_{<k}$ and the whole input of the source sequence $\textbf{e}$ such that:
$
    P(\textbf{s}|\textbf{e};\theta) = \prod_{k=1}^n P(s_k|s_{<k}, \textbf{e};\theta)
$
where $\theta$ are the parameters of the model. A current practice in the community of NLP is to learn these probabilities through Neural Translation (NT) where the tokens are encoded into real vectors.
%

\subsubsection{NT-Architectures considered}
\label{subsubsec:nl-architectures}
We considered three main NT-architectures: sequence to sequence (seq2seq), sequence to sequence plus attention (Att-seq2seq), and the transformer architecture. 

\vspace*{1mm}
\noindent
\textbf{Seq2seq architecture.} 
Seq2seq uses two recurrent neural networks (RNNs), one in the encoder, and one in the decoder, to sequentially process the sentences, word by word~\cite{sutskever2014sequence}.

\vspace*{1mm}
\noindent
\textbf{Att-seq2seq architecture} A drawback of the seq2seq architecture, is that it gradually encodes the dependencies among words in the input and output sentences, by sequentially passing the information to the next cell of the RNN. As a consequence, far-away dependencies may get diluted. In order to correct this problem, an attention mechanism is introduced in att-seq2seq, to explicitly capture and learn these dependencies~\cite{bahdanau2014neural}.

\vspace*{1mm}
\noindent
\textbf{Transformer-architecture} The previous two architectures are relatively slow to train, because the RNNs hinder parallel processing. To alleviate this problem, the transformer architecture, 
introduces a self-attention mechanism, dropping completely the use of RNNs. This dramatically speeded up the computation time of attention-based neural networks, and conferred a considerable momentum to NT~\cite{VaswaniSPUJGKP17}. For this reason we adopted a transformer-based architecture for our DeepSTL translator.

\subsubsection{Preprocessing and Tokenization}
\label{subsubsec:preprocessing}
There are three main features that distinguish our translation problem from general  translation tasks between natural languages (NL2NL): \begin{enumerate}
\item \textit{Out of Vocabulary (OOV)}: The signal names, we call them identifiers for short, and numbers, can be arbitrarily specified. Therefore, it is impossible to maintain a fixed-size vocabulary to cover all imaginable identifiers and numbers. 
\item \textit{High Copying Frequency:} During translation, identifiers and numbers need to be much more frequently copied from the source language to the target language than in NL2NL.
\item \textit{Unbalanced Language:} English is a kind of high-resource language while STL formulas belong to a low-resource logical language that has very limited exclusive vocabulary.
\end{enumerate}

In view of the above characteristics, a successful translation of English to STL requires more than in NL2NL, a correct tokenization of identifiers and numbers. Although one can use an explicit copying mechanism~\cite{copynet2016}, this method requires to modify the structure of the neural network, which may increase complexity.  

\vspace*{1mm}
\noindent
\textbf{Subword tokenization}\quad We therefore adopt a subword technique to tokenize sequences during data preprocessing. Subword algorithms, such as Byte-Pair-Encoding (BPE)~\cite{SennrichHB16a}, WordPiece~\cite{WuSCLNMKCGMKSJL16} and Unigram~\cite{kudo-2018-subword}, are commonly used in state-of-art NT systems to tackle the OOV problem. Without modifying the model structure, these algorithms are able to split words into meaningful morphemes and even independent characters based on statistical rules.

Ideally, we hope when identifiers and numbers are tokenized, they can be respectively encoded by separate characters and digits. For example, \texttt{PWM} and \texttt{12.5} are expected to get encoded as \texttt{[`P', `W', `M']} and \texttt{[`1', `2', `.', `5']} respectively. This way, we can use a limited number of characters and digits to represent arbitrary identifiers and numbers. We chose BPE due to its simplicity.

The tokenization procedure of BPE is executed as follows: (1)~Split every word (separated by a space) in the source data to a sequence of characters. (2)~A prepared token list will include all possible characters (without repetition) in the source data. (3)~The most frequently occurring pair of characters inside a word are merged and added to the token list, then this pair will be treated as an independent character afterwards. (4)~Step 3 is repeated until the size of the token list reaches to an upper limit or a specified hyperparameter. After tokenization, when a sequence is encoded, the generated list is iterated from the longest token to the shortest token attempting to match and substitute substrings for each word in the sequence.

Inspired by BPE algorithm, before tokenization, identifiers and constants have to be split into characters and digits in advance, so that for each pair of two adjacent characters and digits, there would be a whitespace between them (e.g., \texttt{PWM} -> \texttt{P W M}, \texttt{12.5} -> \texttt{1 2 . 5}). In this way, characters and digits will not participate in the merging procedure of BPE.
Hence, during the encoding phase, only characters and digits in the generated token list can match identifiers and constants after they are recognized and split.

During testing time, although it is easy to use regular expressions to match numbers and split them into digits, it is challenging to accurately match identifiers. This is because identifiers can be non-meaningful permutations of characters, or complete English words. These two scenarios cannot be easily distinguished. An ideal method is to adopt Name Entity Recognition (NER) to match identifiers and split them. Now identifiers are automatically identified, by checking that they do not belong to our data-set of commonly used English words, or the English formulation of STL operators.


\subsection{Implementation Details}
\subsubsection{Data split}
We overall generated 120000 English-STL pairs, from which we first sampled 10\% (12000) to prepare a fixed testing set. For the rest, before each training experiment, we sampled 90\% (97200) of them for training, and 10\% (10800) for validation.

\subsubsection{Hyperparameters} 
The implementation of the three models mentioned in~\ref{subsubsec:nl-architectures} are mainly based on ~\cite{zhang2021dive} with several modifications using Pytorch. The following describes how hyperparameters are chosen for each model and the optimizer.

\noindent
\textbf{Seq2seq}\quad 
We used Gated Recurrent Unit (GRU)~\cite{cho2014properties} as RNN units. The encoder is a 2-layer bidirectional RNN with hidden size $h=128$ for each direction, and the decoder is a 2-layer unidirectional RNN with $h=256$. The embedding dimension for mapping a one-hot vector of a token into real valued space is 128. Drop out rate is 0.1.

\noindent
\textbf{Att-seq2seq}\quad For the encoder-decoder, we used the same hyperparameters as Seq2seq-architecture. For Bahdanau attention ~\cite{bahdanau2014neural}, we used a feed-forward neural network with 1 hidden layer and 128 neurons 
to calculate attention score.

\noindent
\textbf{Transformer}\quad For the encoder and the decoder, they both have 4 layers with 8 attention heads; Input and output dimensions for each computing block are always kept as $d_{\text{model}}=128$; Neuron number in feed-forward layers equals to $d_{ff}=512$; Drop out rate is 0.1; For layer normalization, $\epsilon = 10^{-5}$.

\noindent
\textbf{Optimizer}\quad We used Adam Optimization algorithm ~\cite{kingma2014adam} with $\beta_{1}=0.9$, $\beta_{1}=0.98$, $\epsilon=10^{-9}$, while the learning rate $lr$ is dynamically scheduled as (slightly changed from ~\cite{VaswaniSPUJGKP17}):
$$
lr = p \cdot d_{\text{model}}^{-0.5} \cdot \text{min} ( step\_num^{-0.5}, step\_num  \cdot warmup_{steps}^{-1.5})
$$
where $warmup_{steps}=4000$, $d_{\text{model}}=128$. $step\_num$ represents the training steps (training on one batch corresponds to one step). $p$ is an adjustable parameter for each architecture, and we chose 1, 0.1 and 2 for Seq2seq, Att-seq2seq and Transformer respectively. For Seq2seq and Att-seq2seq models, in order to ease gradient explosion due to long sequence dependency, we also used gradient clipping to limit the maximum norm of gradients to 1.

\noindent
\textbf{Other}\quad We dealt with variable length of input and output sequence using padding. We firstly encoded all English and STL sequences into subword token lists, from which we picked the maximum length as the step limit both for the encoder and the decoder. During training, for sequence whose length is smaller than the maximum value, we padded a special token \texttt{<pad>} to its end for complement.

\subsubsection{Train/Validate/Test Procedure} For training and validation, we used ``teacher forcing'' strategy in the decoder. We firstly prepared two special tokens \texttt{<bos>} (begin of sentence) and \texttt{<eos>} (end of sentence). Suppose the reference output of the decoder is \texttt{ABC<eos>}. To start with, we input \texttt{<bos>} as a starting signal to the decoder and hoped that it could output \texttt{A}. No matter whether the actual first output of the decoder is \texttt{A}, we then sent \texttt{A} to the decoder, and hoped that it would output \texttt{B}. This procedure continues until the maximum step length of the decoder is reached. 

We summed up all token-level (only for valid length) cross-entropy loss between the prediction and reference sequence, and divided it by the maximum length of the decoder. This is the loss for one sample. We trained a batch of 64 samples in parallel. The batch loss is averaged over all sample losses inside a single batch, which is used for back propagation to update network parameters.

For testing, ``teacher forcing'' is abandoned. The only token manually input to the decoder is \texttt{<bos>} for initialization. At each time step of decoding, the decoder adopts a greedy search strategy, outputting a token with maximum probability only based on its output in the previous step and the output of the encoder. The decoding procedure will end until the decoder outputs an \texttt{<eos>} token or the maximum limit length is reached.

\subsection{Results}
\subsubsection{Loss/Accuracy Curves}
We trained Seq2seq, Att-seq2seq and Transformer architectures for 80, 10 and 40 epochs respectively, using \emph{STL formula accuracy} (defined in~\ref{subsubsec:test_metrics}) in validation as an indicator to stop training. The validation loss/accuracy curves are obtained by 5 independent experiments and shown as follows.

\begin{figure}[h]
	\includegraphics[width=0.38\textwidth]{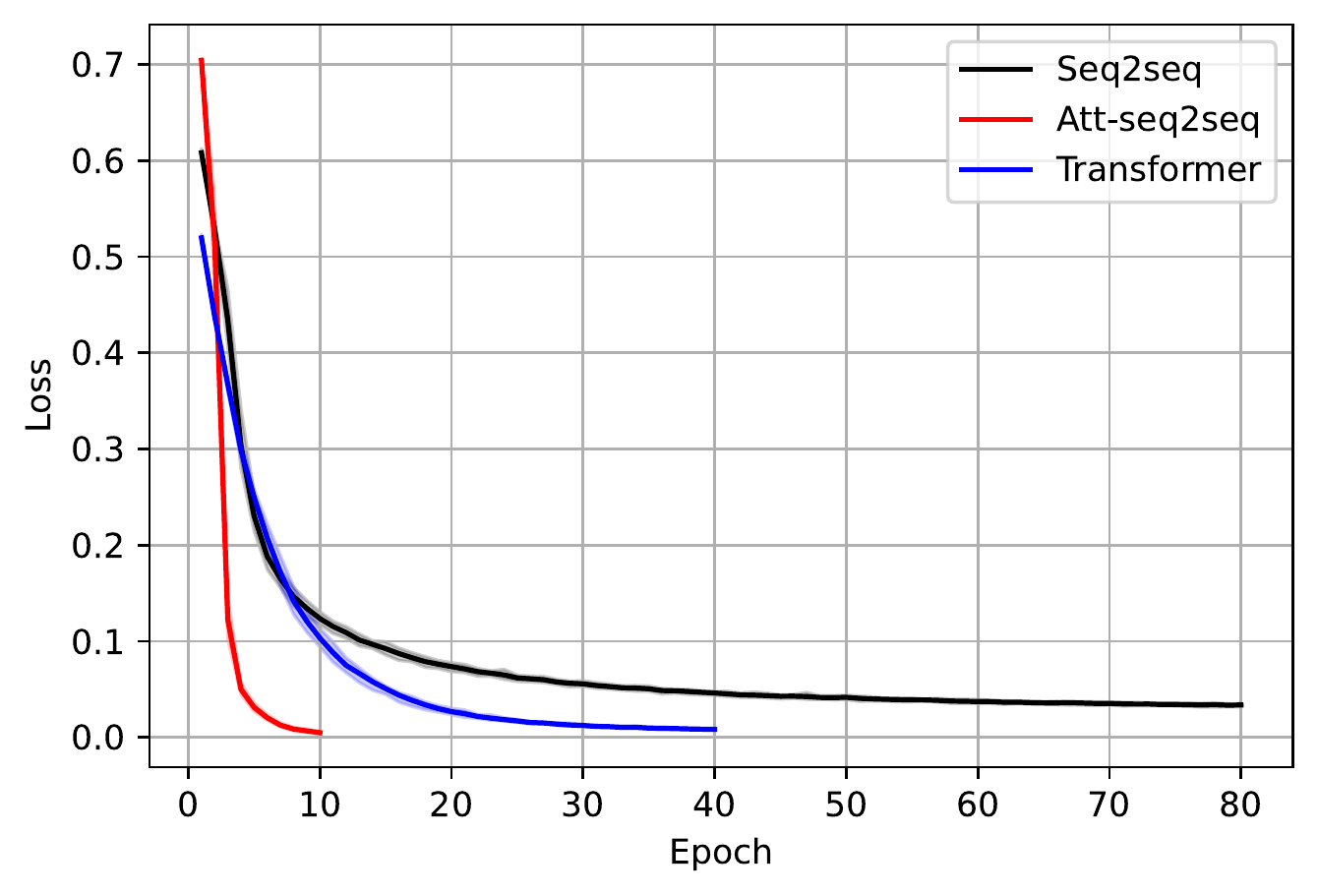}
	\vspace{-3ex}
	\caption{Validation Loss.}
	\vspace{-3ex}
	\label{fig:loss_curve}
\end{figure}

\begin{figure}[h]
	\includegraphics[width=0.38\textwidth]{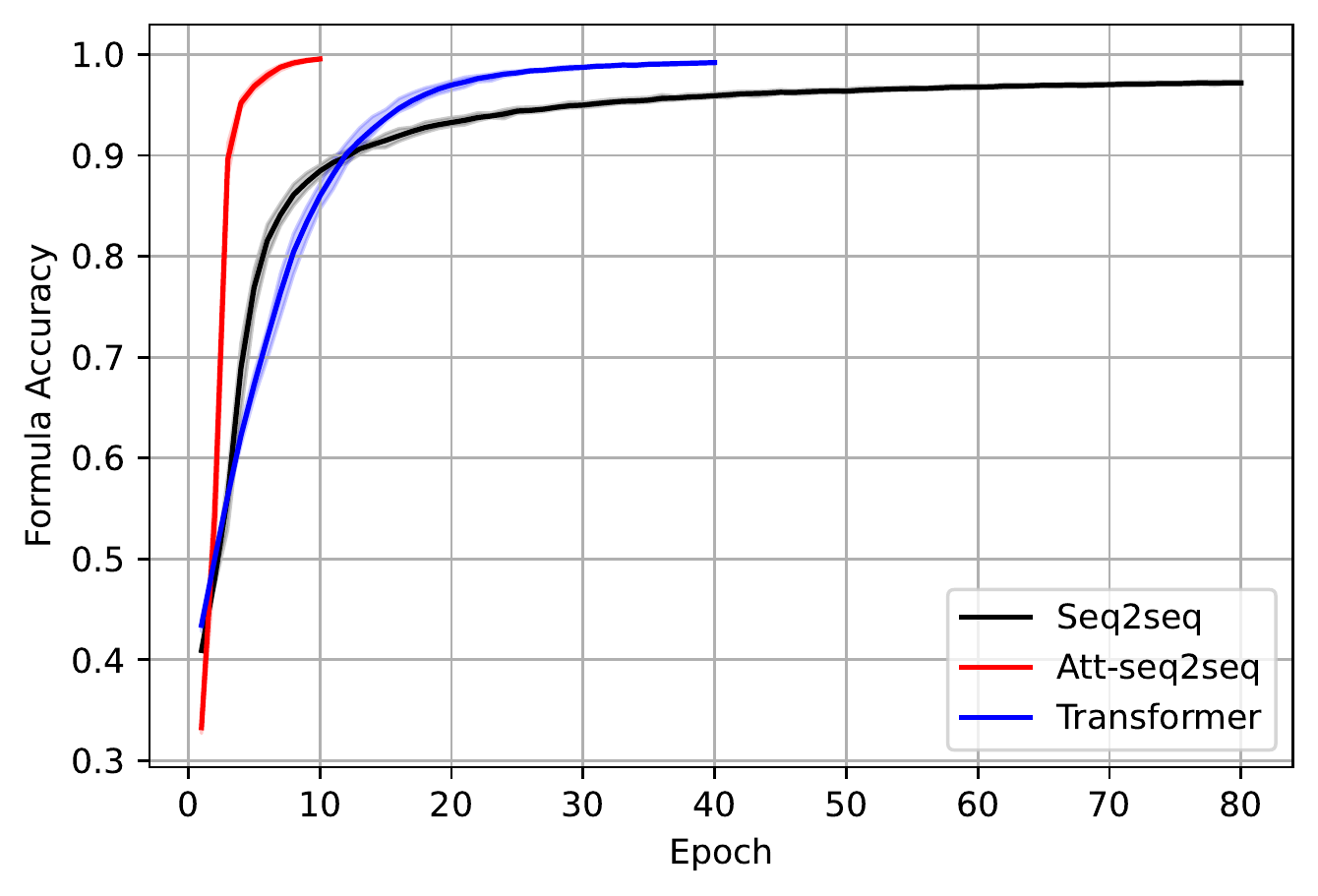}
	\vspace{-3ex}
	\caption{Validation Formula Accuracy.}
	\vspace{-3ex}
	\label{fig:acc_curve}
\end{figure}

Figure~\ref{fig:loss_curve} and Figure~\ref{fig:acc_curve} show that, with the guidance of ``teacher forcing'',  all the three models are able to converge during training, making the STL formula accuracy approach to 1 when the network becomes stabilized. The only difference is the rate of convergence, which depends on many factors like the volume of the model (e.g., number of parameters), noises, learning rate, etc.

\subsubsection{Testing Metrics}
\label{subsubsec:test_metrics}
We firstly report two different measures of accuracy: the \emph{STL formula accuracy} ($A_{F}$)
and the \emph{template accuracy} ($A_{T}$). The first measures the alignment accuracy for the reference and prediction sequence in a string level, while the second firstly transforms the reference and predication instances into STL templates and then calculates their alignment accuracy. For example,
\begin{center}
    Formula: \texttt{always ( x > 0 )} $\ \Rightarrow \ $  Template: \texttt{always ( phi )} \\
    Formula: \texttt{always ( y > 0 )} $\ \Rightarrow \ $  Template: \texttt{always ( phi )}
\end{center}
The first line is reference sequence and the second line represents model prediction. For better illustration, we insert a white-space between each token and thus there are six tokens in the formulas, and four tokens in the templates. For formulas, overall five tokens appear in the same position - \texttt{`always', `(', `>', `0', `)'}, while the left one token \texttt{`x'} in the reference is mistranslated to \texttt{`y'}. Therefore, the formula accuracy $A_{F}=5/6$. As for the template, since all tokens are aligned with each other, the template accuracy $A_{T}=1$.

We also report another metric called BLEU (Bilingual Evaluation Understudy)~\cite{papineni2002bleu} that has been pervasively used in machine translation research. It evaluates the number of n-grams ($n=4$) appearing in the reference sequence. The best BLEU score for a pair of sequences is 1, which means complete overlapping.
\vspace{-2ex}
\begin{table}[h]
\caption{Testing Accuracy.}
\vspace{3ex}
\centering
\begin{tabular}{c|c|c|c} 
 \hline
         & \textbf{Formula Acc.} & \textbf{Template Acc.} & \textbf{BLEU } \\ [0.5ex] 
 \hline
 Seq2seq & $0.071\pm0.0388$ & $0.207\pm0.0868$ & $0.092\pm0.0361$ \\ \hline
 Att-seq2seq & $0.977\pm0.0060$ & $0.980\pm0.0063$ & $0.996\pm0.0011$ \\ \hline
 Transformer & $0.987\pm0.0028$ & $0.995\pm0.0014$ & $0.998\pm0.0005$ \\ \hline
\end{tabular}

\label{table:test_acc}
\end{table}

In Table~\ref{table:test_acc}, it can be seen that once ``teacher forcing'' is removed, the performance of Seq2seq architecture decreases dramatically, which is partly due to its lack of attention mechanism to realize self-correction. For the other two models, both of them can achieve very high accuracy, with Transformer slightly better than Att-seq2seq. Since the testing data and training data are sampled from the same data-set, in this sense, these two models show high translation quality when the distribution of language patterns in testing cases are similar to the training data. We also find that the template accuracy is higher than the formula accuracy. This phenomenon is understandable - once one formula is transformed into the form of template, the potential translation errors in identifiers, constants and logical relation symbols are masked.

\subsubsection{Extrapolation}
\label{subsubsec:extrapolate}
In the following, we use the informal requirements that we identified from the literature in Section~\ref{sec:empirical_stl} to evaluate how well the 
machine learning algorithm generalizes the translation 
outside of the training and validation data set.


In order to have a fair evaluation, we used the 
$14$ \textbf{Clear} requirements ($10\%$ of the 
entire set) with the template 
structure supported by our tool. We pre-processed 
the requirements to remove units that are not supported 
by our tool. Table~\ref{table:extra_acc} summarizes 
the accuracy results for the three learning approaches. 
We see that with non-synthetic examples the formula 
accuracy drops considerably for all algorithms, while 
the average template accuracy remains relatively high ($89.9\%$) 
for the Transformer approach. We believe that higher 
availability of publicly available informal requirements 
that could be used for training would considerably help 
improving the accuracy of the approach.

\vspace{-2ex}
\begin{table}[h]
\caption{Extrapolation Accuracy.}
\vspace{3ex}
\centering
\begin{tabular}{c|c|c|c} 
 \hline
         & \textbf{Formula Acc.} & \textbf{Template Acc.} & \textbf{BLEU } \\ [0.5ex] 
 \hline
 Seq2seq & $0.050\pm0.0283$ & $0.158\pm0.0895$ & $0.027\pm0.0120$ \\ \hline
 Att-seq2seq & $0.559\pm0.0865$ & $0.742\pm0.0660$ & $0.888\pm0.0348$ \\ \hline
 Transformer & $0.712\pm0.0678$ & $0.899\pm0.0100$ & $0.962\pm0.0030$ \\ \hline
\end{tabular}

\label{table:extra_acc}
\end{table}


In the following, we provide three examples \footnote{These examples are the actual outputs of the translator. They are not displayed in a mathematical way. The part that is incorrectly translated is represented in blue color.} that illustrate the possibilities and the limits of our approach (random seed $=100$). We also report the following metric that considers the average logarithmic value of the output confidence at each decoding step:
$
    C_m = \frac{1}{L^\alpha} \sum_{k=1}^L \text{log} \ P(s_k|s_{<k}, \textbf{e};\theta)
$
where $L$ is the length of the output sequence, $\alpha$ is an adjustable factor which is set to $0.75$ by default, and $m \in \{ s, a, t \}$ with $s$, $a$, $t$ denoting Seq2seq, Att-seq2seq and Transformer model respectively.


\vspace*{1mm}
\noindent \textbf{Example 1:} If the value of signal RWs\_angular\_momentum is greater than 0.35, then the value of signal RWs\_torque shall be equal to 0.~\cite{boufaied2021signal}
\begin{itemize}
    \item \textbf{Transformer ($C_t = -0.01393$)}: \\
    always ( RWs\_angular\_momentum > 0.35 -> RWs\_torque == 0 )
    \item \textbf{Att-seq2seq ($C_a = -0.30038$)}: \\
    always ( RWs\_angular\_m{\color{blue}xyomemeEqm < 0.3} -> RWs\_torque==0)
    \item \textbf{Seq2seq ($C_s = -2.77145$)}: \\
    always ( {\color{blue} WNcAi1iDSDDyD1yD2y171a71aa2345324621 ) 5} {\color{red} ...... too long, display omitted}
\end{itemize}


\vspace*{1mm}
\noindent \textbf{Example 2:} Whenever Op\_Cmd changes to Passive then in response Spd\_Act changes to 0 after at most 500 time units.
\begin{itemize}
    \item \textbf{Transformer ($C_t = -0.00091$)}: \\
    always ( rise ( Op\_Cmd == Passive ) -> eventually [ 0 : 500 ] ( {\color{blue} rise} ( Spd\_Act == 0 ) ) )
    \item \textbf{Att-seq2seq ($C_a = -0.10360$)}: \\
   always ( rise ( Op\_Cmd == Passive ) -> {\color{blue} not} ( {eventually} [ 0 : 500 ] ( Spd\_Act == 0 ) ) )
    \item \textbf{Seq2seq ($C_s = -3.03260$)}: \\
    always ( rise ( {\color{blue} PIweD > 12.3 Q8y5yDy6y1y1R11y1y1g1y1A} {\color{red} ...... too long, display omitted}
\end{itemize}

\noindent \textbf{Example 3:} Whenever V\_Mot enters the range [1, 12] then in response starting after at most 100 time units Spd\_Act must be in the range [100, 1000].
\begin{itemize}
    \item \textbf{Transformer ($C_t = -0.00873$)}: \\
    always ( rise ( V\_Mot >= 1 and V\_Mot <= 12 ) -> eventually [ 0 : 100 ] ( Spd\_Act >= 100 and Spd\_Act <= 1000 ) )
    \item \textbf{Att-seq2seq ($C_a = -0.06080$)}: \\
    always ( rise ( V\_Mot >= 1 and V\_Mot <= 12 ) -> {\color{blue} not} ( eventually [ 0 : 100 ] ( Spd\_Act >= 100 and Spd\_Act <= 1000 ) ) )
    \item \textbf{Seq2seq ($C_s = -2.68981$)}: \\
    always ( rise ( {\color{blue} p\_qHX > 4 Q3DaQaDamymaOlQ ) ya ) 4 fall} {\color{red} ...... too long, display omitted}

\end{itemize}

The extrapolation test shows the poor translation of 
Seq2seq that is consistent with its low accuracy
measured in~Table ~\ref{table:test_acc}. The translation
quality of Transformer and Att-seq2seq is much higher. 
It is however sensitive to how similar the 
patterns used in the informal requirement are to 
the ones used in the training data. 

In Example 1, 
Transformer makes the correct translation, while 
Att-seq2seq fails to copy the identifier and the number, and ``greater than'' is mistranslated into ``$<$''. 
%
%
%
In Example 2, Transformer tends to add a \rise ~ operator before the subformula wrapped inside an $\F$ operator, although in some occasions this is equivalent to the actual intention of the requirement because ``changes to'' often indicates the significance of a rising edge. On the other side, Att-seq2seq adds a negation operator $\neg$ in front of the subformula starting with an $\F$ operator, so the meaning becomes reversed. In Example 3, Transformer translates the requirement correctly while Att-seq2seq makes the same mistake. For the three examples considered, the Seq2seq model fails to translate all of them, without even guessing correctly the template: it tends to generate lengthy symbols without explicit meaning.

%% file: 7_discussion.tex
\section{Discussion}
\label{sec:discussion}

\noindent
\textbf{Corpus Generator}\quad
One of the major limitations is that the natural language is still generated through a rule-based approach with human intervention. Although for commonly-used STL formulas, the corpus generator can already produce enormous fluent synonymous translations, the diversity in expression is still limited. However, it is this ``cold-start'' approach that makes it possible in the future to adopt automatic data augmentation techniques~\cite{feng2021survey} in NLP to produce much more English utterances exponentially. These new variants may involve different linguistic features, such as ambiguity and vagueness.

Furthermore, the English requirements produced from the corpus generator are generic texts strictly following the semantics of STL without incorporating terminology and domain knowledge of a particular field, e.g., electronics, robotics, and biology. How to cover, characterize and process this information with modular design patterns is a future research direction.
\\\\
\noindent
\textbf{Neural Translator}\quad
An important improvement would be to unify the pipeline in data preprocessing: we need to combine Name Entity Recognition (NER) ~\cite{li2020survey, yadav2019survey} technique with the look-up table method to recognize arbitrarily designated identifiers, or those that are already given in signal tables from industrial data sheets.

Besides, translation accuracy and decoding confidence should be further exploited for training because they are indicators of translation performance. In fact, the template accuracy mentioned in \ref{subsubsec:test_metrics} is biased by design in that it penalizes positional mismatches (with cumulative effect) more strongly than individual token mismatching errors. Hence, an unbiased criterion for quantifying template accuracy needs to be considered. For translation confidence, a high level is not necessarily indicating that the decoder will always insist on the correct translation - sometimes it implies that the decoder may be stubborn to the wrong output (Example 2 of Transformer in \ref{subsubsec:extrapolate}). However, for lower confidence values, they tend to indicate insufficient training of a particular feature due to unbalanced training samples. Given the rich information conveyed, these two metrics can be promisingly used as feedback signals to guide the optimization of the loss function so that different problems occurring in training can be detected and corrected.

Furthermore, due to the attention mechanism, Att-seq2seq is still a strong competitor to Transformer, despite being inferior in certain test metrics and cases. Although it is interesting to improve translation quality from multiple Transformer-based pre-training techniques ~\cite{qiu2020pre} with large models, it is also important to figure out what role attention mechanism exactly plays in translation: unlike the English in daily communications or in literature, the contexts used to specify formal languages like STL are relatively limited.
Thus, an English-STL translator that only depends from statistical information learned from a data set  (as generally it occurs in NL2NL translators) may be not ideal.
 
However, the approach presented here still depends on learning the statistical features of the training data rather than really understanding the requirement language. This is the reason 
for the mistranslation of ``greater than'' into ``$<$'' in
Example 1 of Att-seq2seq. 
Given this, in \ref{subsubsec:extrapolate} it is reasonable to see the drop in accuracy when testing cases are very different from what have been trained. To build DeepSTL with enhanced trustworthiness, the integration of syntactic and semantic information into the end-to-end learning process using attention mechanism will be a topic of further investigation.

Finally, from a user interaction point of view, the next generation of DeepSTL should output multiple possible translations for the user to choose from, thus remembering the user's language preferences in order to provide customized service.

%% file: 8_conclusion.tex
\section{Conclusion}
\label{sec:conclusion}

We studied the problem of translating CPS natural language requirements to STL, commonly 
used to formally specify CPS properties by the academic community and 
practitioners.
To address the lack of publicly available 
natural language requirements, we developed a procedure 
for automatically generating English sentences from STL formulas. 
We employed a transformer-based NLP architecture to efficiently train an accurate translator 
from English to STL. Experiments demonstrated
promising results.

While this work focuses on STL specifications 
and CPS applications, the underlying principles can be applied to other domains and specification formalisms and have a significant positive impact on the field of requirement engineering. 
Unlike natural languages, 
formal specifications have a very 
constrained structure. We believe that this observation can be further explored in the future to develop an even more robust translation mechanism and thus further strengthen requirements engineering methodologies.